\documentclass[a4paper]{llncs}

\usepackage{geometry}
\usepackage{amsmath}
\usepackage{amsfonts}
\usepackage{multirow}
\usepackage{graphicx}
\usepackage{subcaption}
\usepackage{fancyhdr}
\usepackage[ruled,vlined]{algorithm2e}
\usepackage[sorting=none]{biblatex}
\newcommand*\samethanks[1][\value{footnote}]{\footnotemark[#1]}
\DeclareUnicodeCharacter{0301}{é}

\graphicspath{{images/}}

\fancypagestyle{paper}{
    \fancyhf{} 
    \fancyfoot[c]{\thepage} 
}
\bibliography{references}
\begin{document}

\title{Neural Architecture Search in operational context: a remote sensing case-study}

\author{Anthony Cazasnoves \thanks{Authors contributed equally to this paper} \and
Pierre-Antoine Ganaye \samethanks \and
Kévin Sanchis \samethanks \and 
Tugdual Ceillier}

\institute{Preligens (ex-Earthcube), Paris, France \\ \url{www.preligens.com}, [name].[surname]@preligens.com}
\maketitle              

\begin{abstract}
Deep learning has become in recent years a cornerstone tool fueling key innovations in the industry, such as autonomous driving. To attain good performances, the neural network architecture used for a given application must be chosen with care. These architectures are often handcrafted and therefore prone to human biases and sub-optimal selection. Neural Architecture Search (NAS) is a framework introduced to mitigate such risks by jointly optimizing the network architectures and its weights. Albeit its novelty, it was applied on complex tasks with significant results -- e.g. semantic image segmentation. In this technical paper, we aim to evaluate its ability to tackle a challenging operational task: semantic segmentation of objects of interest in satellite imagery. Designing a NAS framework is not trivial and has strong dependencies to hardware constraints. We therefore motivate our NAS approach selection and provide corresponding implementation details. We also present novel ideas to carry out other such use-case studies.
\end{abstract}

\keywords{Neural Architecture Search (NAS) \and High Performance Computing (HPC) \and Remote Sensing \and Semantic segmentation.}


\section{Introduction}
\paragraph{}Deep learning has nowadays established itself as the reference framework to tackle complex signal processing problems. Their high level of performance and strong versatility make neural networks (NN) both best contenders on many public benchmarks and key enablers for multiple real-world applications, among which autonomous driving, expert translation system, 3D pointcloud processing or remote sensing. Reaching operational performances through deep learning is nonetheless not trivial. The quantity of high-quality labelled data is obviously one of the process cornerstones. Another one lies in the architecture of the network that will be trained.

\paragraph{}In terms of architecture design, one can think of two potentially antagonist objectives to reach: finding the best performing structure possible while enforcing a complexity level suitable for training. This trade-off performance / simplicity is perceptible in the literature. When considering a specific task, some deceptively simple structures manage to reach very good performance levels and further gains can only be made at considerable complexity increase costs. Compare, for example, the UNet\cite{unet} to the DeepLab V3\cite{chen2017rethinking} when dealing with semantic segmentation. In most contributions, the network structure is purely handcrafted, either inspired by other \textit{NN} or distilled know-how from non machine-learning algorithms. Many atypical architectures are, in this way, never probed: human design is subject to multiple bias -- e.g. preference for symmetries in structures -- that can lead to the exclusion of simple architecture exhibiting good performances.
\paragraph{}Automated Machine-Learning (Auto-ML)\cite{yao2019taking} is the most encompassing framework to tackle such concerns; the idea there being to model both the algorithm task and all your possible choices for the network architecture as optimization problems. Under such paradigm, these critical selections are performed through a much more complex gradient descent. Neural Architecture Search (NAS) is a sub-domain of Auto-ML: it consists in jointly optimizing the NN weights and its architecture to find the most suitable solution to a given problem -- all other hyper-parameters remaining hand picked. While mainly applied on \textit{lower complexity} tasks -- e.g. image classification with large inter-classes variance, such as CIFAR-10\cite{cifar10} or ImageNet\cite{deng2009imagenet} -- NAS has recently shown encouraging results on tasks closer to real-world use-cases. AutoDeepLab \cite{autodeeplab} for example managed to outperform SOTA results on the CityScape benchmark\cite{cityscapes}, thus proposing an interesting architecture for autonomous-driving applications.

\paragraph{}In this technical paper, we evaluate the potential of NAS for one of our company use-cases: the semantic segmentation of objects of interest in satellite imagery. To the best of our knowledge, NAS was never applied in this operating context. We have 3 objectives. First, we aim to evaluate if this technology can prove useful in finding relevant network architectures for remote sensing applications. Considering the higher level of complexity involved in implementing a NAS framework, we also strive to provide a detailed technical feedback relevant to people interested in developing one. Finally, we intend to highlight the limitations encountered and introduce some ideas for further studies.

\paragraph{}This paper is organized as follows. Section \ref{sec:bibliography} first introduces the NAS problem formulation and the two main families of approaches used to perform such research. Selection of relevant methods is then motivated in Section \ref{sec:selected_methods}. Experiments taxonomy is presented in Section \ref{sec:experiments} while results are detailed in Section \ref{sec:results}. Conclusions and discussions of this study can be found in Section \ref{sec:conclusions}. Implementation challenges and related caveats are available in Appendix \ref{sec:implementation}. 


\section{Problem statement and related works}\label{sec:bibliography}

\paragraph{}In this section, we'll briefly introduce how the NAS problem is often framed in literature before presenting the two main families of approaches to conduct said search.

\subsection{Problem statement: an overview}

\paragraph{}We here follow the insight developed in \cite{elsken2019neural, wistuba2019survey} and therefore encourage interested readers to consult these surveys for more in-depth details.
\paragraph{}As aforementioned, the bread and butter of NAS is to automatically probe for the best architecture to tackle a specific problem. In supervised learning, a problem is mainly modeled through an annotated dataset. As such, letting $\mathcal{D}$ be the space of all annotated datasets, $\mathcal{A}$ the search space of NN architectures and $\mathcal{M}$ the trained models one, the NAS search application $\Gamma$ is defined under 
\begin{equation}
\Gamma: \mathcal{D} \times \mathcal{A} \rightarrow \mathcal{M}.
\end{equation}

\noindent
In deep learning, upon selecting a specific architecture $a \in \mathcal{A}$ and addressing the problem considered $d = (d_{train}, d_{valid}) \in \mathcal{D}^2$, one wants to find the best set of network weights $w$ by minimizing

\begin{equation}
\Gamma(a, d_{train}) = \underset{m_{a, w}\in \mathcal{M}}{{\arg\min}} \mathcal{L}(m_{a, w}, d_{train} ) + \mathcal{R}(w),
\end{equation}

\noindent where $m_{a, w}$ represents the trained model, $\mathcal{L}$ the selected loss function and $\mathcal{R}$ the regularization term.

\noindent Building upon this, NAS strives to find the best architecture $a^*$ to handle problem $d$ i.e. the one searched on $d_{train}$ exhibiting best performance on $d_{valid}$ according to the objective function $\mathcal{I}$:

\begin{equation}
\label{eq:problem}
a^* = \underset{a \in \mathcal{A}}{{\arg\max}} \:  \mathcal{I}(\Gamma(a, d_{train}), d_{valid}).
\end{equation}

\begin{figure}[!ht]
\centering
  \includegraphics[width=0.95\textwidth]{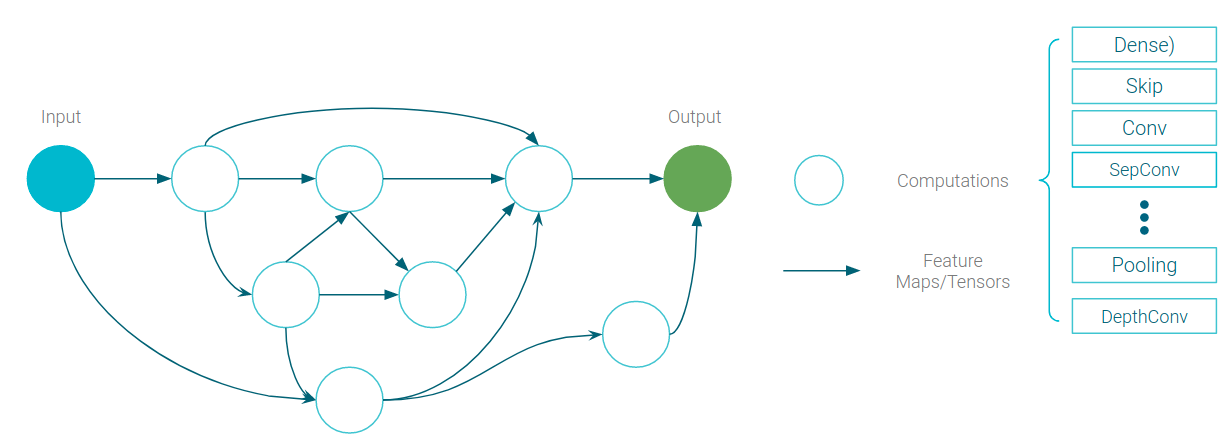}
  \caption{A NN can be symbolically encoded using two structures: a directed acyclical graph (DAG) for its topology (left) and a dictionary of operations available to perform computations (right).}
  \label{img:dag_topo}
\end{figure}

\noindent $\mathcal{I}$ being a black-box function, no direct optimization can be performed. Two families of methods have been proposed to get around this problem, it will be the topic of the following sections.
\paragraph{}Before that, some concepts common to both families must be introduced. Enabling an automated search of architecture requires to define and model the search space $\mathcal{A}$ embedding all possible architectures. One can for example choose to encode it as two complementary structures -- see Figure~\ref{img:dag_topo}. Firstly, a directed acyclic graph (DAG) that can encode all available networks topologies by:
\begin{itemize}
    \item mapping tensors to its nodes
    \item mapping operations to its edges
    \item letting the graph direction follow the computation ones
\end{itemize}

\noindent Secondly, the set of all available operations $\mathcal{O}$ that can be selected to perform said computations. In doing so, the search strategy can then select any topologies available in the DAG, pick any operations available in $\mathcal{O}$ and instantiate a NN that will be trained and evaluated.

\subsection{Discrete Formalism}

\paragraph{}The discrete approaches were historically the first propositions in the NAS community. These methods proceed in an exploratory way, that could be summed-up in three main phases:

\begin{enumerate}
    \item choose a subset of NN (by picking topologies and related operations)
    \item train this subset (classical deep learning)
    \item evaluate it according to relevant criteria
\end{enumerate}

\noindent
As a first naive approach, one can of course iterate over these three steps by picking at random the probed subsets and recording the history of resulting performances to later choose its best performer. However, considering the combinatorial number of possible architectures, the time required to find a good contender for a given real-world problem -- without supervision in this search -- will be considerable. As we do have access to a relevant evaluation metric, it can be used to supervise the search. Two branches of algorithms are well suited for such a task: evolutionary algorithms (EA) and reinforcement learning (RL).
\paragraph{}From a very general perspective, RL is based on a feedback loop mechanism. An agent performs actions affecting its environment, which impacts are then evaluated by a controller that provides the adequate guiding signal for the update policy. Iterating in this way, the agent adapts and improves its performances -- i.e. its ability to correctly behave on the training task . Following this idea in NAS, one can choose to set a controller acting as architecture sampler, guided by its achieved performance. In \cite{zoph2016neural} a RNN acts as the controller that provides a variable-length token encoding the NN architecture. The NN performances are used as rewards driving the RNN -- either through REINFORCE loss \cite{zoph2016neural} or Proximal Policy Optimization \cite{zoph2018learning}. Other contributions chose to exploit the Q-Learning paradigm to perform this sampling \cite{baker2016designing, zhong2018practical}.  

\paragraph{}EA strives to emulate a nature-inspired adaptation processed, as framed in Darwinian theory. Using a specific fitness criterion, and a population of potential solutions, the algorithm loops through the following steps to select the best candidates:

\begin{enumerate}
    \item best specimens selection (information selection)
    \item matting (information sharing)
    \item mutating (random information perturbation)
\end{enumerate}

\noindent There is a large variety of ways to perform these steps and to handle specific inherent risks like sub-optimal elitist selection. In NAS, the population will be composed of architectures genotypes -- structures encoded as strings, for instance. The evaluation will be performed on their phenotypes, i.e. NN that have been instantiated and trained. In terms of mutations, two types can be explored jointly: operands switch and topological modifications. \cite{liu2018hierarchical} uses tournament selection while \cite{real2019regularized} builds upon it by introducing a new parameter: ageing. In a running population, a fraction of the oldest individuals is removed at each cycle -- indiscriminately from their performance -- to limit the risk of early stagnation and provide a wider exploration breadth. An interesting proposal is introduced in \cite{elsken2019efficient}: maintaining an architecture population by sampling the Pareto front performances / resources requirement. It is shown that along generations, the achieved trade-offs are improving significantly. 

\paragraph{}All the techniques cited above suffer from the same major limitation: no matter how smart the exploration strategy, trying to discover $a^*$ in this way amounts to an almost brute-force exploration of all available architectures. One can take advantage of an HPC infrastructure to improve the chances: at each iteration, the number of candidates will scale in agreement with the number of GPUs available. To have a real impact, however, the resources involved are colossal -- as seen in many papers, they are on the scale of hundreds to thousands of GPUs. Recent studies \cite{wistuba2019survey, liu2021survey, yu2019evaluating} even point out that EA and RL may not outperform random search. These drawbacks have led to the development of another paradigm: the idea that a better problem formulation could partially alleviate the hardware constraint and provide a greater chance of success in finding a good architecture.

\subsection{Continuous Formalism}\label{sec:continuous}

\paragraph{}Building on the idea of simplifying the training process and reducing the significant computation cost of aforementioned methods, a new set of approaches based on the training of a single, dense network for architecture search were proposed. Herein, we review the main methods of this group and define the formalism which they are based upon.

\paragraph{}DARTS \cite{darts} is the first method to consider architecture search as the training of a single differentiable hyper-network that contains the whole operation search space, using the principle of continuous relaxation. We review this method as the core building block of continuous approaches in architecture search. As mentioned previously, $\mathcal{O}$ is our operation search space, a set of predefined operations that can be applied to a tensor $x$, such as a convolution or a pooling. DARTS aims at building a search network $\mathcal{S}$, that is a directed acyclic graph (DAG) formed of a single search cell that is used several times. This search cell is in fact another DAG: training $\mathcal{S}$ amounts to assigning an operation ${o_{i}}$ from $\mathcal{O}$ to each edge of this sub graph.\\
Formally, during training, each edge of the search cell is assigned to a mixed operation ${o_{m}}$ defined as a weighted sum over all operations from $\mathcal{O}$:

\begin{equation}
\label{eqn:continuous_relaxation}
o_{m}^{(i, j)}(x)=\sum_{o \in \mathcal{O}} \frac{\exp \left(\alpha_{o}^{(i, j)}\right)}{\sum_{o^{\prime} \in \mathcal{O}} \exp \left(\alpha_{o^{\prime}}^{(i, j)}\right)} o(x)
\end{equation}
\newline

\noindent Where \( \alpha_{o}^{(i, j)} \) is a vector of dimension \( \vert{\mathcal{O}}\vert \) associated with an edge \((i, j)\), that is jointly learned alongside operation weights \(w\) with gradient descent.\\
Note that, as an edge \( (i, j) \) connects two nodes \( n_i \) and \( n_j \), a node's output can be expressed as follows:
\newline

\begin{equation}
\label{eqn:darts_node}
{n_j}=\sum_{i<j} o_m^{(i, j)}
\end{equation}
\newline

\noindent
Once training of \(w\) and \( \alpha_{o} \) is done, a discrete architecture can be decoded from \( \mathcal{S} \) by replacing each mixed operation \( {o_{m}}\) with the best operation \( o^* \) w.r.t \( \alpha \), defined as: 
\newline

\begin{equation}
\label{eqn:darts_decode}
o^{*(i, j)}=\operatorname{argmax}_{o \in \mathcal{O}} \alpha_{o}^{(i, j)}
\end{equation}
\newline

 \paragraph{}Building on the concept of differentiable architecture search, several methods were proposed afterwards, which resulted in a new family of approaches based on continuous relaxation, as opposed to the more traditional discrete ones.

 \paragraph{}\cite{dartsplus} proposed an incremental improvement to solve a well identified problem in DARTS-based approaches, where sometimes skip or any other operation that contains few or no parameters would be highly preferred over operations with parameters (e.g convolutions), leading to poor performance once the decoded discrete network is retrained. This work suggests to use manually defined criteria to control the number of skip connections in the discretized search cell. \cite{pcdarts} tackled the issue of memory efficiency induced by DARTS-based approaches, inherent to the construction of a super-network that encapsulates the whole operation search space (e.g mixed operations). By using channel-wise partially connected operations, this method manages to significantly reduce the memory footprint of the search network. Additionally, PC-DARTS introduces edge normalization, a new form of regularization that helps overcoming the \textit{skip-connection problem} described above.\cite{autodeeplab} goes a step further by learning and decoding both the network structure and the cell structure simultaneously. It proceeds by building a densely connected search network -- as opposed to cell-based networks traditionally used by DARTS approaches -- that is also trained using continuous relaxation and additional $\beta$ parameters encoding the network connection probabilities between cells. Finally, the Viterbi algorithm is used to decode a network structure, as in this case decoding an optimal structure from a densely connected DAG means finding an optimal path from the network's start to its end.
 

\section{Selected methods}\label{sec:selected_methods}

\paragraph{}Considering the dedicated literature, multiple methods can be used to tackle our use-case through NAS. We will here introduce the reasons behind our methods and implementation choices.

\subsection{Choosing the most appropriate paradigm}

\paragraph{}As mentioned before, Neural Architecture Search algorithms can be separated into two categories: discrete search and continuous relaxation of the search space. Although we explored both paradigms, we determined that one was more adapted to our use case. While exploring the typical discrete framework by training multiple architectures using a simple genetic algorithm, we encountered several challenges, the most prominent one being the unpredictable memory footprint and training time of such approaches. More specifically, we could not anticipate the GPU RAM usage of the sampled architectures, leading to uncontrolled program failures due to memory allocation limit\cite{gao2020estimating}.

\paragraph{}Considering this observation, we explored the more stable continuous approaches and deemed them adapted to our use-case. We first considered the classical DARTS approach. While this method offers a significant reduction in memory and training time, it is constructed on a cell-based network and tailored to the image classification problem. In order to build suitable architectures for our image segmentation task, we replaced the usual cell-based structure by another one, similar to U-Net \cite{unet}. Undoubtedly, this new structure is not as lightweight as the original one, and thus lead to an increase in memory consumption. However, this consumption is constant throughout training and methods like the ones used for PC-DARTS~\cite{pcdarts} can help reducing it.

\subsection{Reducing the memory footprint}

\paragraph{}In the purpose of dealing with the memory issue described in the previous section, we explored \cite{pcdarts} as a potential solution. \textit{Partial channel connection}~\cite{pcdarts} is an effective technique to reduce the memory overhead of computing mixed operations inside every search cell. For a given edge \( (i, j) \) containing a mixed operation \( o_{m}^{(i,j)}\), the input tensor \(x\) is split into two parts along the channel axis. The computation is only applied on the first part, while the second part is simply propagated and left unchanged using a skip connection. Formally, a mask \(M_{i,j}\) assigns 1 to channels inside the first part, and 0 to the others. Thus, \( o_{m}^{(i,j)}\) is defined as:
\newline

\begin{equation}
\label{eqn:pc_update}
o_{m}^{(i, j)}(x;M_{i, j})=\sum_{o \in \mathcal{O}} \frac{\exp \left(\alpha_{o}^{(i, j)}\right)}{\sum_{o^{\prime} \in \mathcal{O}} \exp \left(\alpha_{o^{\prime}}^{(i, j)}\right)} o\left(M_{i, j} * x_{i}\right)+\left(1-M_{i, j}\right) * x_{i}
\end{equation}
\newline

\noindent
The weighted sum is only applied on a subset of channels -- as opposed to Equation~\ref{eqn:continuous_relaxation} where every operation is applied on $x$ -- which reduces the overall memory footprint. Practically, the mask is applied by randomly sampling a portion of 1 / \(K\) channels that will be processed -- \(K\) being an hyper-parameter. This solution creates an adjustable trade-off between efficiency, when \(K\) is large, and accuracy, when \(K\) is small. Additionally, sampling a subset of channels adds a form of regularization to the architecture search, as most channels are skipped and the influence of \( \alpha \) on the overall operation selection is reduced. However, random sampling also adds instability to the training process.

\paragraph{}We follow \cite{pcdarts} and use \textit{edge normalization} in conjunction with partial channel connection. Formally, edge normalization is defined as the addition of a value, denoted \( \gamma^{(i,k)} \), that is bound to each edge of a search cell, so that the expression of a node \(n_j\) (\ref{eqn:darts_node}) becomes:
\newline

\begin{equation}
\label{eqn:edge_normalization_node}
{n_j}=\sum_{i<j} \frac{\exp \left(\gamma^{(i, j)}\right)}{\sum_{i^{\prime}<j} \exp \left(\gamma^{(i^{\prime}, j)}\right)} o_m^{(i, j)}
\end{equation}
\newline

\noindent
and the decoding rule (\ref{eqn:darts_decode}) is now expressed as:
\newline

\begin{equation}
\label{eqn:edge_normalization_decode}
o^{*(i, j)}=\operatorname{argmax}_{o \in \mathcal{O}} \alpha_{o}^{(i, j)} \gamma^{(i,j)}
\end{equation}
\newline

\noindent
Similar to \( \alpha \), \( \gamma \) are learned throughout the training and bring stability to counteract the random, unstable nature of channel sampling. Albeit the use of partial channel connection and edge normalization solved the memory footprint overhead inherent to DARTS-based approaches, we empirically observed an additional issue bound to the operation selection step, detailed in the following section.

\subsection{Improving architecture selection}
\begin{figure}[!ht]
  \includegraphics[width=1\textwidth]{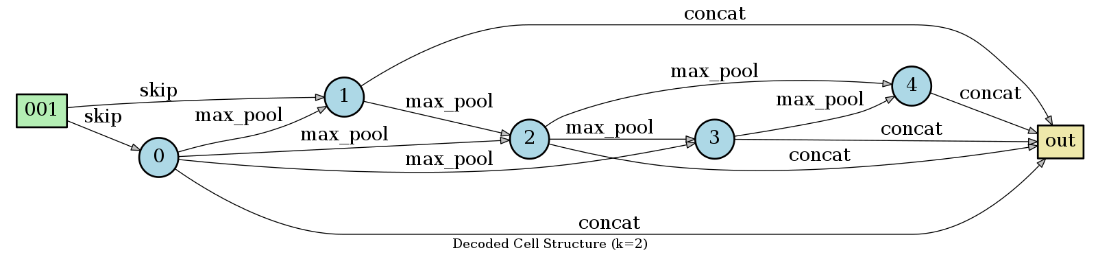}
  \caption{Graph of a decoded (post-search) DARTS cell. Only operations without learnable parameters were selected. This type of weight-free architecture is called degenerated because of its lack of modeling capacity.}
  \label{img:degenerated_darts_cell}
\end{figure}

\paragraph{}While the continuous relaxation scheme has gained a lot of interest for its efficiency, we experimentally observed several failure cases: poor operations diversity within the decoded cell (see Figure \ref{img:degenerated_darts_cell}) and performance discrepancy between the search step and the retraining steps. These problems have already been mentioned in section \ref{sec:selected_methods} and in \cite{nasproblems}.

\paragraph{}To correct the generalization issues, \cite{gaea} formulated an optimization method called GAEA (Geometry Aware Exponentiated Algorithm), which is more suitable than Stochastic Gradient Descent (SGD) for updating the architecture's parameters. Indeed, GAEA is based on exponentiated gradient descent which is known to make convergence faster for constrained parameters and ultimately provides sparsity properties that are helpful to reduce the generalization gap induced by DARTS \cite{darts}. In the optimization method proposed in \cite{gaea}, the network's parameters are still learned by SGD, while GAEAis used to optimize only \( \Theta \), the architecture's weights.

\paragraph{}GAEA can be plugged into continuous relaxation-based approaches such as PC-DARTS by substituting the learned architecture parameters $\alpha_{o}^{(i, j)}$ by $\theta_{o}^{(i, j)} \in \Theta=\mathbb{R}^{|\mathcal{O}|}$.
The main difference between $\alpha_{o}^{(i, j)}$ and $\theta_{o}^{(i, j)}$ is that $\theta_{o}^{(i, j)}$ is constrained to be a probability vector (probability of operation $o \in \mathcal{O}$ for edge$(i,j)$), where $\alpha_{o}^{(i, j)}$ is only considered as a logit that is used by the softmax layer.
Optimizing $\theta_{o}^{(i, j)}$ consists in two steps: updating with exponentiated gradient (equation \ref{eqn:gaea_update}), then re-parameterizing $\theta_{o}^{(i, j)}$ (equation \ref{eqn:gaea_reparam}):

\begin{equation}
\label{eqn:gaea_update}
\theta^{\prime (i, j)} \leftarrow \theta^{(i, j)} \odot \exp \left(-\eta \nabla_{\theta} f\left(\mathbf{w}, \theta^{(i, j)} \right)\right)
\end{equation}

\begin{equation}
\label{eqn:gaea_reparam}
\theta_{o}^{\prime (i, j)} \leftarrow \frac{\theta_{o}^{\prime (i, j) }}{\sum_{o^{\prime} \in O} \theta_{o^{\prime}}^{\prime (i, j)}} \forall o \in O
\end{equation}

\noindent
where $\eta$ is the learning rate and $\theta_{o}^{\prime (i, j)}$ is the probability of operation $o$ being selected for edge $(i,j)$ at the next parameter update.

\begin{figure}[!ht]
  \includegraphics[width=1\textwidth]{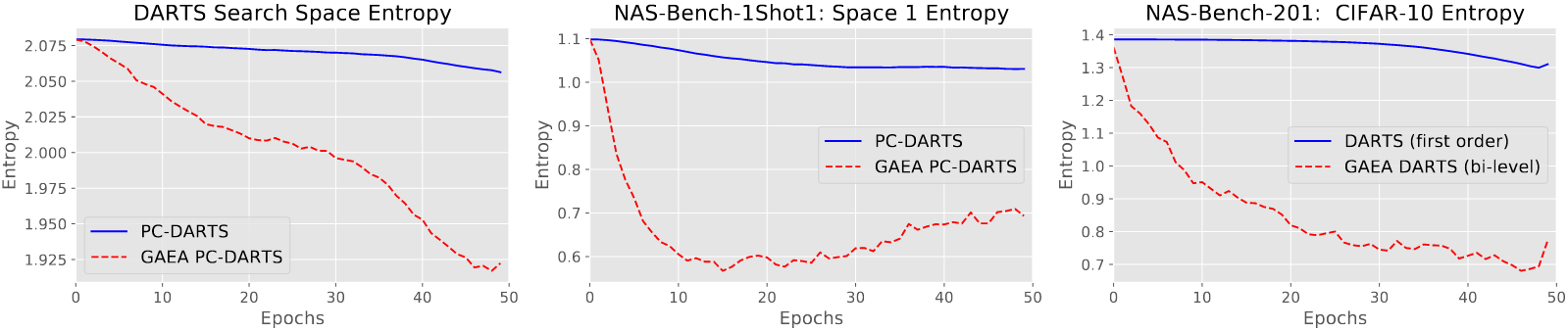}
  \caption{Evolution of the entropy of $\theta$ during architecture search for PC-DARTS \cite{pcdarts} and GAEA \cite{gaea}. Figure from \cite{gaea}.}
  \label{img:gaea_entropy}
\end{figure}

\paragraph{}One way to measure the progression of architecture search is to use the entropy of $\theta$ (the lower, the more certain the choice of operation $o$ is). The active learning community uses entropy as a measure of uncertainty. Here, it can be applied on $\theta$ to derive a convergence indicator for the architecture's parameters. As mentioned previously, the main advantage of GAEA is its improved convergence speed, which translates into an increased $\theta$ sparsity and thus lower entropy. When compared to PC-DARTS, we can observe in Figure~\ref{img:gaea_entropy} that GAEA reaches a lower entropy by a significant margin.

\section{Experiments}\label{sec:experiments}
\subsection{Use-case presentation}

\noindent
\paragraph{}We perform neural architecture search for a semantic segmentation task. Our dataset contains around 23,000 patches with 12,000 observables, tiled from \textit{Maxar/DigitalGlobe} satellites WorldView 1, 2, and 3 and GeoEye 1 publicly available full-size images. The patches have a dimension of 512\( \times \)512 pixels, with an overlap of 180 pixels between adjacent patches. We use 18,000 patches for training and 5,000 patches for validation. We do not follow the common 50/50 split of the training dataset used by DARTS \cite{darts} for optimizing layer weights and architecture weights, but instead follow the method used by GAEA \cite{gaea}, which consists in duplicating and using the full training dataset for both the layer weights and the architecture weights. Finally, we evaluate our decoded architectures on a testing set composed of 38 full-size images, containing 906 observables. The objective of our task is to assign one of two classes to each pixel: \textit{background} or \textit{helicopter}.

\begin{figure}[!ht]
    \centering
    \includegraphics[scale=0.5]{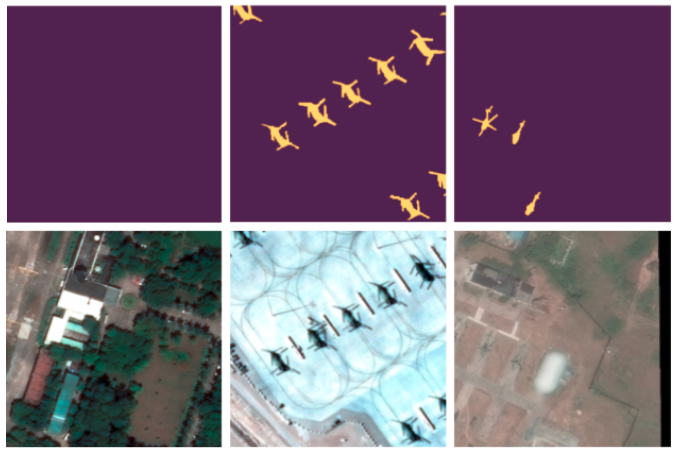}
    \caption{Typical examples from our dataset. Helicopters can be found at any position and have different forms. Some images have a significant class imbalance or contain no objects. Sometimes, helicopters can be hard to distinguish from the background.}
    \label{fig:dataset_examples}
\end{figure}

\paragraph{}Remote sensing images bring various challenges. First, there is usually a significant class imbalance between background and objects, as objects are relatively small. Objects also don't follow any placing rules and can be found at any position on a patch. Finally, tiling an image also creates corner cases where objects are at the edge of a patch, sometimes partially visible, because of cuts between two patches. Additionally, in our specific case, objects can have different shapes, as illustrated in Figure~\ref{fig:dataset_examples}. To solve this task, we explore different search spaces, network and cell structures thoroughly described in the following section.

\subsection{Explored topologies}

\paragraph{}We present all conducted experiments sorted by \textit{topology}. We define a topology \( \mathcal{T} \) as a unique combination of network structure \(\mathcal{S}_n\), cell structure \(\mathcal{S}_c\), and operation search space \( \mathcal{O} \). Presented topologies adopt the following naming convention: \(\mathcal{S}_c\)-\(\mathcal{S}_n\). To perform these experiments, we developed an internal NAS framework that aims to be as generic as possible. Further details about this framework are available in \textbf{Appendix \ref{sec:implementation}}.

\paragraph{}We first review the \textbf{ResNeXt-UNet} topology. In this setting, we use a network structure identical to U-Net \cite{unet}, more fitted for instance segmentation than the classical cell-based structure of DARTS. Furthermore, the DARTS cell is discarded in favor of a structure similar to a ResNeXt \cite{resnext} block. The reasoning behind the choice of this topology is to restrain our search network \( \mathcal{S} \) to follow already proven structures on the task we tackle. In other words, we apply NAS on a well-defined setting and try to incrementally improve upon an already convenient structure, as well as evaluating the method on a simpler case (i.e with less degrees of freedom) than DARTS.

\begin{figure}[!ht]
    \centering
    \includegraphics[scale=0.25]{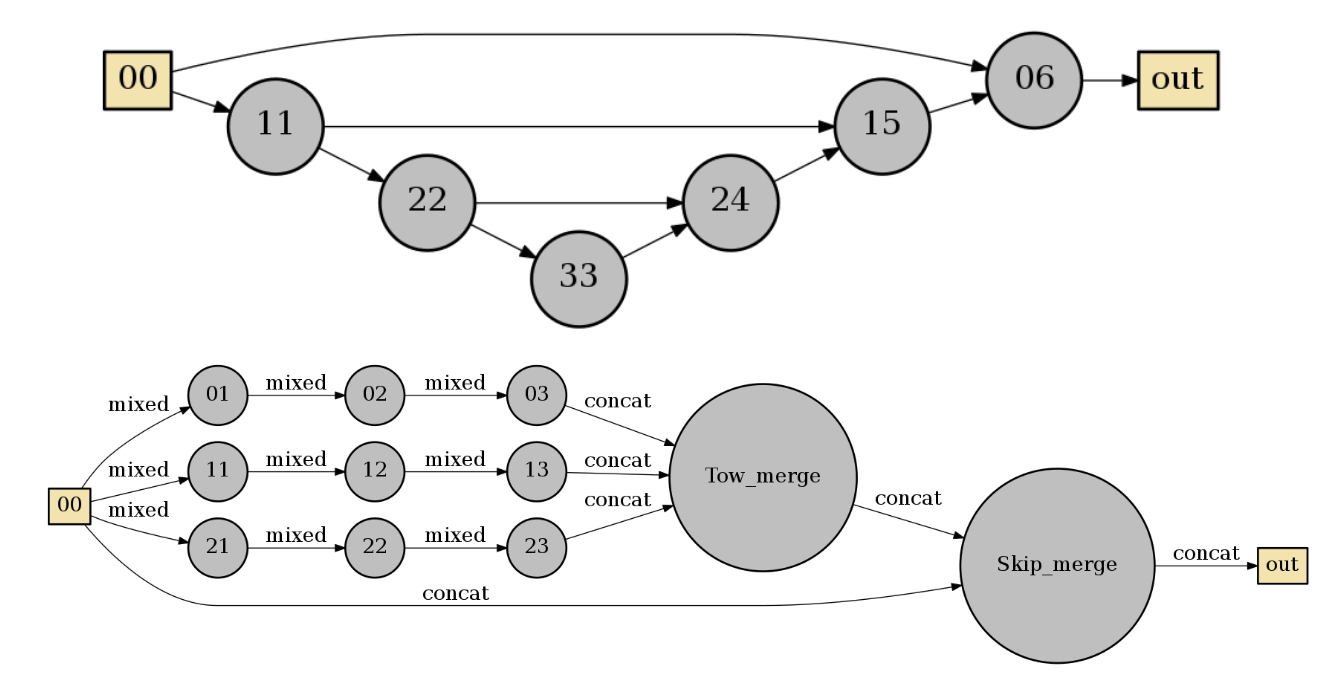}
    \caption{The ResNeXt-UNet topology. The upper graph is the search network, the lower graph is the search cell.}
    \label{fig:resnext_unet}
\end{figure}

\paragraph{}In \cite{darts,pcdarts,dartsplus}, the usual operation search space that we denote \( \mathcal{O}_{base} \) contains the following operations: 3\(\times\)3 and 5\(\times\)5 dilated convolutions, 3\(\times\)3 and 5\(\times\)5 depthwise separable convolutions, 3\(\times\)3 average pooling, 3\(\times\)3 max pooling, skip connection, and zero (e.g multiplying the input tensor by 0). To compensate for the lower degrees of freedom, we design a larger operation search space \( \mathcal{O}_{large} \) containing:  3\(\times\)3, 5\(\times\)5 and 7\(\times\)7 convolutions, 3\(\times\)3, 5\(\times\)5 and 7\(\times\)7 depthwise convolutions, 3\(\times\)3, 5\(\times\)5 and 7\(\times\)7  spatial separable convolutions, skip connection, and zero. In our implementation, we respectively call the operations \textit{conv2d\_1}, \textit{conv2d\_2}, \textit{conv2d\_3}, \textit{depthconv2d\_1}, \textit{depthconv2d\_2}, \textit{depthconv2d\_3}, \textit{splitconv2d\_1}, \textit{splitconv2d\_2}, \textit{splitconv2d\_3}, \textit{skip}, and \textit{cut}. Essentially, we replaced pooling operations with normal convolutions, and added a 7\(\times\)7 kernel size for all convolutions. The ResNeXt-UNet topology is depicted in Figure~\ref{fig:resnext_unet}.

\begin{figure}[!ht]
\centering
\begin{subfigure}{.4\textwidth}
  \centering
  \includegraphics[width=0.95\linewidth]{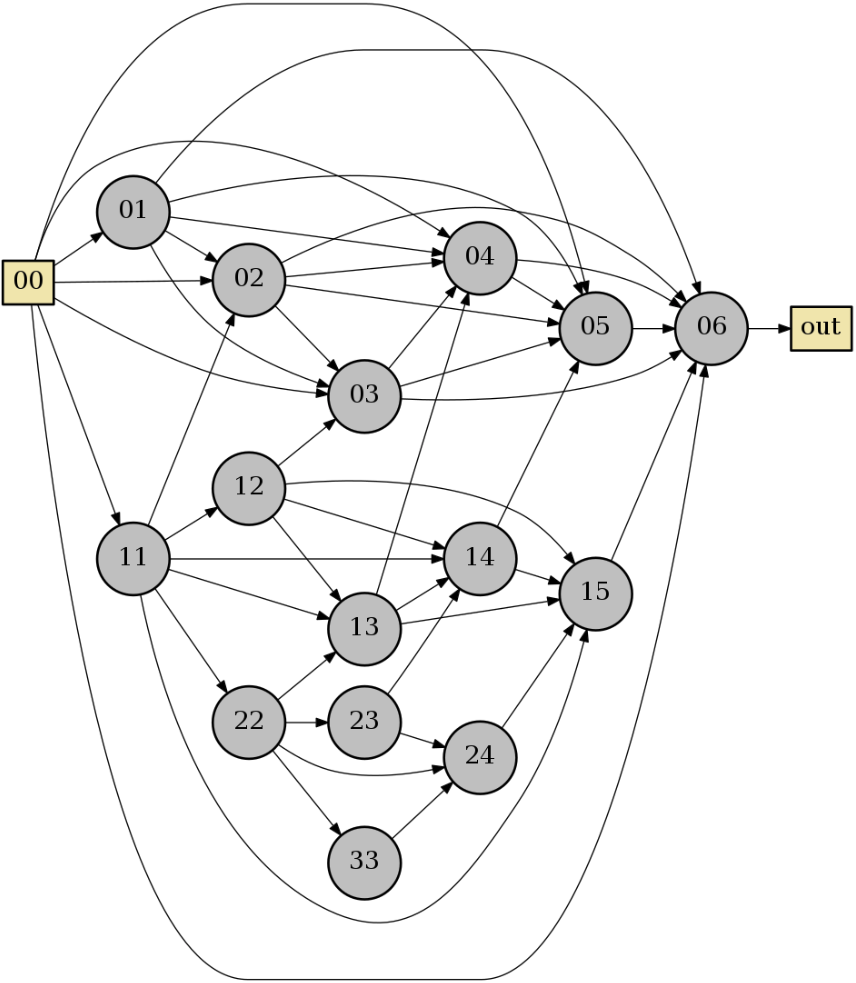}
  \caption{The UNet++ network topology.}
  \label{fig:unetpp_network}
\end{subfigure}%
\begin{subfigure}{.60\textwidth}
  \centering
  \includegraphics[width=0.9\linewidth]{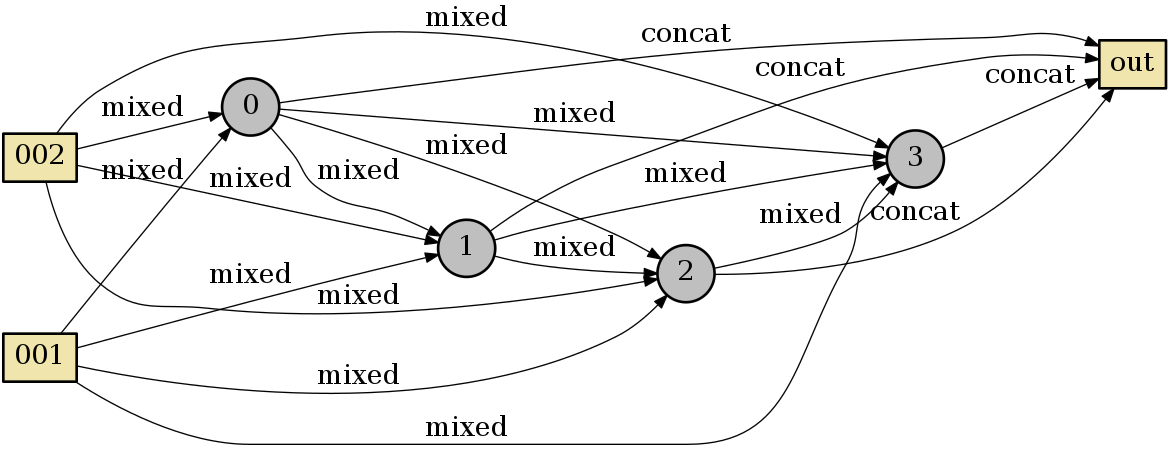}
  \caption{The DARTS cell topology. As in cell-based networks, this search cell is connected to its parent node [001] and grand-parent node [002].}
  \label{fig:darts_cell}
\end{subfigure}
\caption{Topologies of the UNet++ network (a) and the DARTS cell (b)}
\label{fig:cell_topologies}
\end{figure}

\paragraph{}Concurrently, we explore the \textbf{DARTS-UNet} topology, the closest to a typical DARTS-based topology. Building on the previous intuition, we use the same U-Net structure that we associate with a classical DARTS cell, depicted in Figure~\ref{fig:darts_cell}. Similarly to \cite{darts,pcdarts,dartsplus}, our search cell contains 4 blocks, and we use \( \mathcal{O}_{base} \) as an operation search space. We also follow the same connection scheme as a cell-based network structure, where each search cell is connected to its parent node (i.e the closest first order neighbour in topological order), and its grand-parent node (i.e the closest second order neighbour in topological order).

\paragraph{}While decoding the trained DARTS cell, we keep the \( \mathcal{K} \) strongest input edges per block based on:

\begin{equation}
\label{eqn:darts_cell_decode}
\alpha_{o^*}^{(i,j)}=\operatorname{max}_{o \in \mathcal{O}} \alpha_{o}^{(i, j)}
\end{equation}
\newline

\noindent
and discard the others. In other words, we sort edges using the \(\alpha\) value associated with their selected operation and retain the \(\mathcal{K}\) largest values for each block. Akin to \cite{darts,pcdarts,dartsplus}, we set \( \mathcal{K} \) to 2, so each block is connected to at most 2 input edges after decoding.

\paragraph{}Using this topology, we aim at providing an adaptation of DARTS-based approaches for an instance segmentation task, only replacing the network structure. To the best of our knowledge, there is no existing work that propose such a direct application of DARTS on image segmentation. Albeit some methods such as \cite{nasunet,autodeeplab} tackle this task, they require significant changes and are not straight applications of DARTS. Building on the idea of a two-step decoding from \cite{autodeeplab}, we additionally consider a denser network structure in the \textbf{DARTS-UNet++} topology. Based on \cite{unet++}, this structure remains close to U-Net, but is also dense enough to contain novel encoder-decoder structures.

\paragraph{}In \cite{autodeeplab}, the cell-based network from DARTS is replaced with a densely connected one that is trained similarly to search cells, i.e with \( \beta \) continuous values associated with all edges. After training, a single path is decoded using the Viterbi Algorithm, where softmaxed \( \beta \) values are considered as transition probabilities. This optimal path is then used as an encoder and attached to a fixed decoder. We extend this idea to train a dense encoder-decoder search network. We define a network structure identical to \cite{unet++}, as shown in Figure~\ref{fig:unetpp_network}, and aim at discovering an encoder-decoder structure. We associate this network structure with DARTS search cells. As we are looking for an encoder-decoder network, decoding a single path is not suitable. Instead, we propose a simple, greedy multi-path algorithm that computes the average log-likelihood of all possible paths. 
While this algorithm is not optimized, the set of paths is limited enough -- e.g. lesser than 100 -- to remain computationally tractable. Upon gathering their scores, one still needs to select a relevant subset of $\mathcal{K}$ paths used in the decoded architecture. $\mathcal{K}$ may either be considered as a fixed hyper-parameter, or one can try to build a statistical rule to decide the adequate cardinal to retain. We attempted a simple tryout that exhibited convincing properties in our experiments. Letting $p_{sc} \in \mathbb{R}^\vert p_{sc} \vert$ be the set of paths scores, one can compute its corresponding mean $\mu$ and variance $\sigma$. Assuming Gaussian nature, $\mathcal{K}$ may therefore be determined as:
\begin{equation}
    \mathcal{K} = \vert {p_{sc}, p_{sc} \geq \mu + 3*\sigma} \vert
\end{equation}

\noindent
We describe the results of our experiments in the next section.

\section{Results and observations}\label{sec:results}

\paragraph{}In this section, we present and discuss the results of our experiments. We compare the performance of decoded and retrained search networks with a custom \textit{baseline network} following a ResNet-UNet architecture -- a U-Net where encoder blocks are replaced with ResNet \cite{resnet} blocks, and tailored for our dataset. In other words, hyperparameters such as the number of filters of each convolutional layer, or the number of layers in each block, are handpicked to achieve the best possible results on our dataset. We report in the following sections the \textit{Precision-Recall curves} on the testing set and evaluate our topologies against the baseline network. 

\paragraph{}For all experiments described in the following subsections, similar to \cite{darts}, we use an initial number of channels of 16 and start training \( \alpha \) values at epoch 15, using GAEA \cite{gaea} as an optimizer for \( \alpha \), and SGD with momentum for layer weights \(w\). Furthermore, we use a cosine annealing learning rate with an initial value of 0.01, down to 0 for SGD, and a fixed learning rate of 0.1 for GAEA. Finally, we apply a weight decay of \(10^{-3}\). Furthermore, similar to the search step, we use SGD with the cosine annealing rate mentioned above as an optimizer for the training of the decoded architecture. Finally, we use weighted cross entropy as a loss for all experiments.

\subsection{ResNeXt-UNet}\label{sec:resnext_unet}

\paragraph{}We first investigate the results of the ResNeXt-UNet topology. We train the search network for 50 epochs, using a batch size of 2, a partial channel connection of 1/8, and no edge normalization as it is not applicable for a ResNeXt cell, since each block is only connected to at most one input edge. Once a network architecture is decoded, we train it for 80 epochs and select the checkpoint that achieves the best \textit{MeanIoU} on the validation set, starting from the 30th epoch.

\begin{figure}[!ht]
    \centering
    \includegraphics[scale=0.20]{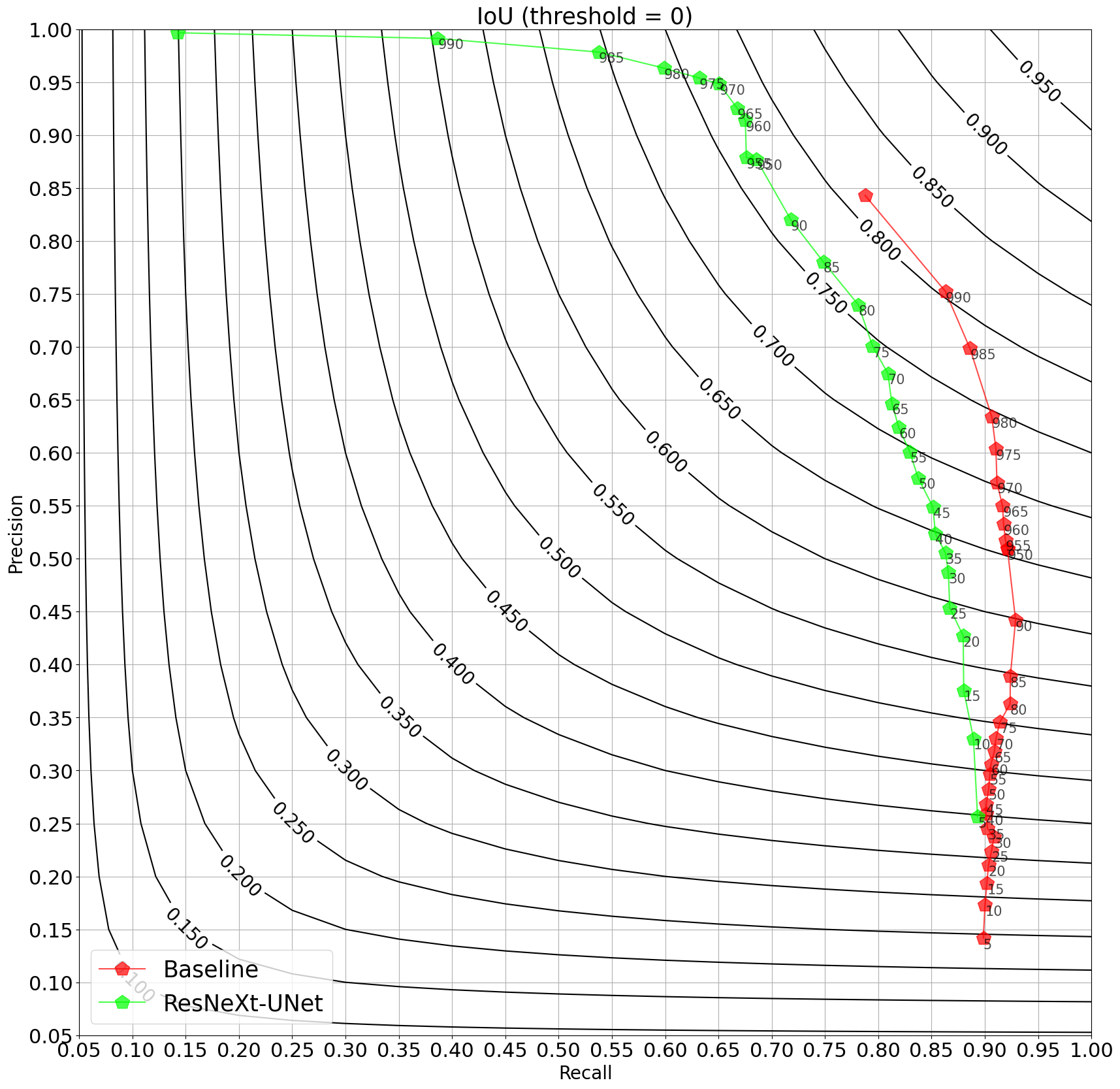}
    \caption{Precision-Recall values as a function of IoU thresholds.}
    \label{fig:resnext_rps}
\end{figure}

\paragraph{}We report the achieved \textit{Precision-Recall curve} in Figure~\ref{fig:resnext_rps}. Although we did not manage to do better than the baseline network, we emphasize the relatively small difference between both approaches and compare time and efforts required to elaborate them. As the baseline network is manually designed and optimized for our dataset, it requires a significant amount of time and efforts to create, whereas our ResNeXt-UNet only required a single architecture search. Additionally, we did not increase the number of filters or add any form of regularization in the decoded ResNeXt-UNet, albeit the wide usage of such a trick in DARTS-based approaches \cite{darts,pcdarts,dartsplus,autodeeplab}. Finally, we did not use any form of data augmentation while retraining our decoded network, which is not true for the baseline.

\begin{figure}[!ht]
    \centering
    \includegraphics[scale=0.32]{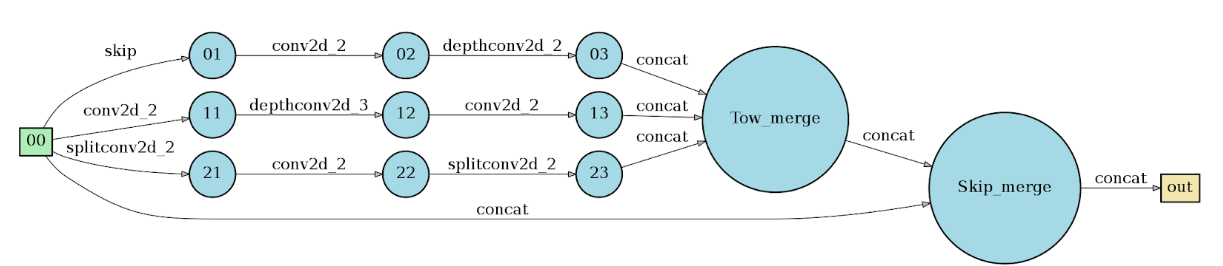}
    \caption{The decoded ResNeXt cell (K=2).}
    \label{fig:decoded_resnext}
\end{figure}

\paragraph{}In Figure~\ref{fig:decoded_resnext}, we show the decoded ResNeXt cell. We notice the majority of selected convolutions uses a \( 5\times5 \) kernel, except \textit{depthconv2d\_3} which uses a \( 7\times7 \) kernel. Furthermore, selected operations contain a good variety of convolutions, as all types are chosen twice, and only one skip connection, indicating that the training successfully avoided the \textit{skip connection problem}. Finally, each tower branch is different -- which may seem counter-intuitive -- and contains no real logic. It is safe to assume it can be difficult for a human to manually design this architecture.

\begin{figure}[!ht]
    \centering
    \includegraphics[scale=0.4]{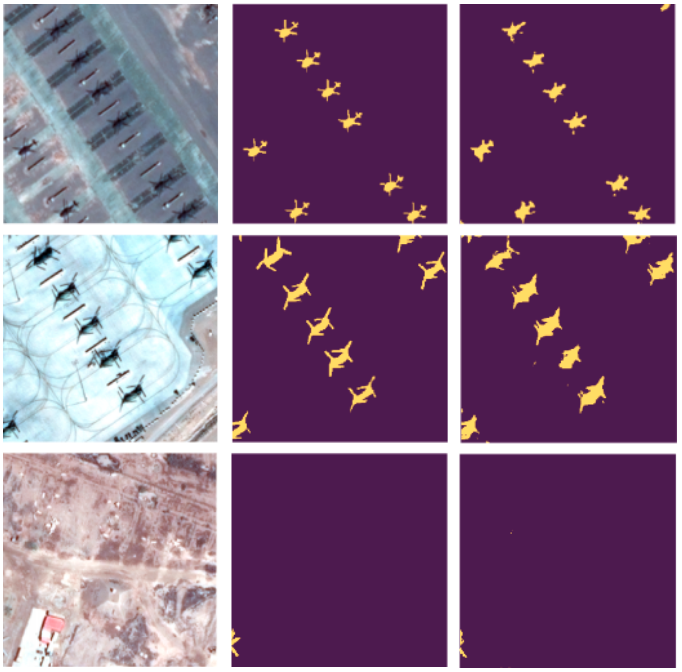}
    \caption{Predictions on the testing set. \textbf{(left)} The ground truth. \textbf{(right)} The prediction.}
    \label{fig:resnext_preds}
\end{figure}

\paragraph{}We present a small sample of predictions on the testing set in Figure~\ref{fig:resnext_preds}. Globally, our network makes few false positives, but has a hard time contouring objects with precision, although being able to detect the general shape. 

\subsection{DARTS-UNet}

\paragraph{}In DARTS \cite{darts}, the authors experiment on image classification with two types of cells, namely the normal and reduction ones. Both share a common search space, the main difference being that the stride of the first nodes within the reduction cell is set to 2 (instead of 1), in order to apply a spatial reduction of the resolution. In the DARTS-UNet experiment, we follow the same scheme by designing a UNet \cite{unet} architecture with reduction cells within the encoder and normal cells with the decoder part.

\paragraph{}The DARTS-UNet search space follows the one proposed in \cite{pcdarts}, it is only composed of pooling layers, separable and dilated convolutions and skip connection. Because these operations are light in terms of parameters, we can increase the batch size from 2 to 4, compared to other experiments. The training of the best architecture is done with SGD for 200 epochs, as we observed a slower convergence compared to the ResNeXt-UNet architecture.

\begin{figure}[!ht]
\centering
\begin{subfigure}{.4\textwidth}
  \centering
  \includegraphics[width=0.95\linewidth]{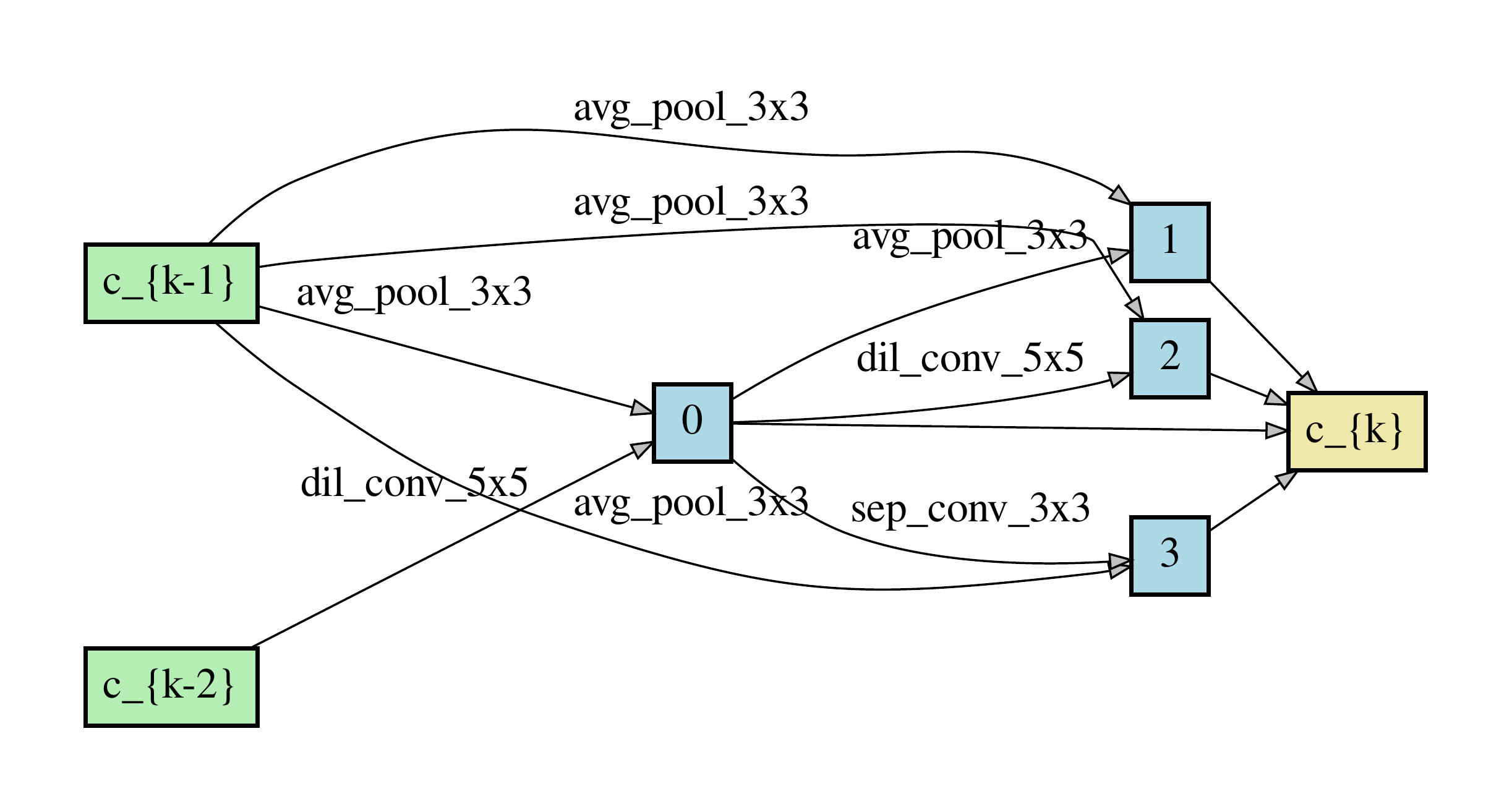}
  \caption{DARTS-UNet normal cell decoded after search.}
\end{subfigure}%
\begin{subfigure}{.60\textwidth}
  \centering
  \includegraphics[width=0.9\linewidth]{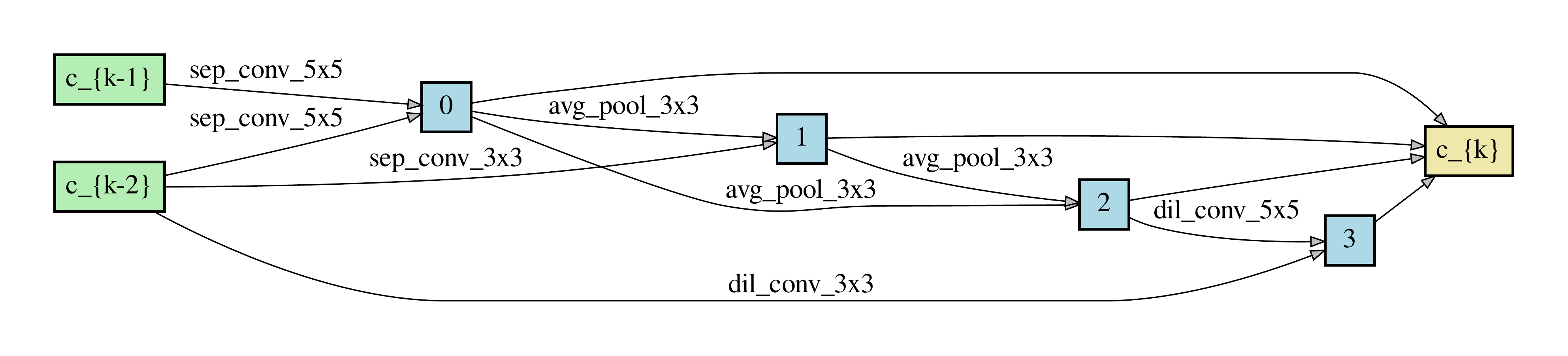}
  \caption{DARTS-UNet reduction cell decoded after search}
\end{subfigure}
\caption{DARTS-UNet cells obtained after the architecture search.}
\label{fig:darts_unet}
\end{figure}

\paragraph{}During our experiments, the maximum IoU for the search step was 0.52, while the training of the sampled architecture led to an IoU of 0.73. The cell structures obtained after the search step can be visualised in figure \ref{fig:darts_unet}. We notice the heavy presence of average poolings in both normal and decoded cells, and the absence of skip connections and max poolings. Furthermore, a good variety of convolutions in terms of type and kernel size can also be noticed. Once again, the absence of skip connections may indicate that the training did not fall into the \textit{skip connection problem}, despite preferring operations with a low amount of parameters (e.g. pooling). The fair amount of convolutions seems to indicate a successful training nonetheless. Finally, we notice a lower amount of average pooling inside the reduction cell, which seems intuitive since this type of cell already reduces the number of parameters by dividing the input feature map by two, and thus does not need much pooling on top of it.

\subsection{DARTS-UNet++}
\paragraph{}For this topology, the search step is performed on 50 epochs, and the decoded architecture is trained on 80 epochs. As main differences, edge normalization is here applied and we use two stem-cells -- fixed cells used to reduce the input feature map's size at the start of the network -- before entering the main DARTS-UNet++ architecture -- thus reducing the spatial resolution to $128\times128$ pixels. This stemming scheme is mainly due to VRAM consumption limitation and enables us to perform a wider search in the main architecture section, as the DARTS-UNet++ is significantly heavier than other architectures in terms of memory requirements.

\begin{figure}[!ht]
\centering
\begin{subfigure}{.45\textwidth}
  \centering
  \includegraphics[width=0.95\linewidth]{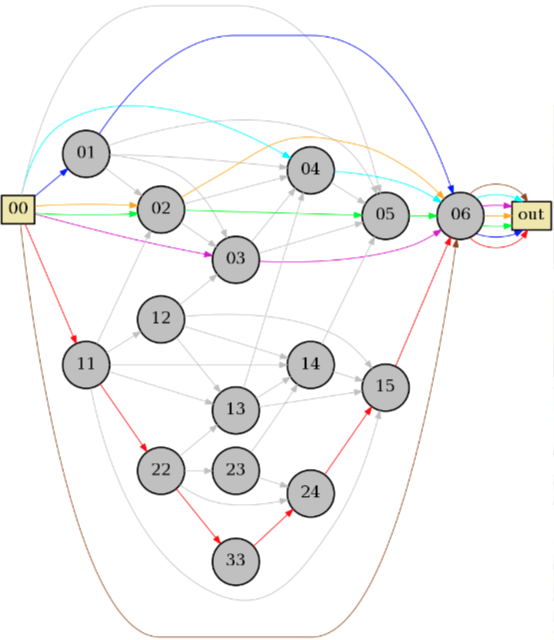}
\end{subfigure}%
\begin{subfigure}{.55\textwidth}
  \centering
  \includegraphics[width=0.95\linewidth]{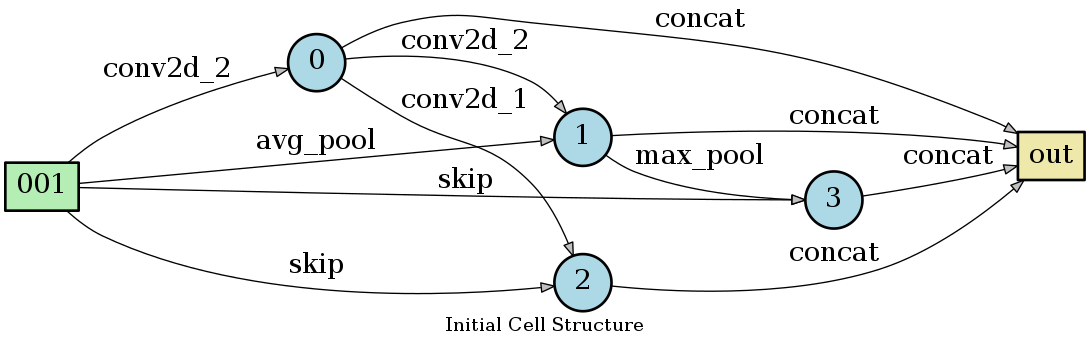}
\end{subfigure}
\caption{Decoded UNet++ network (left, colored paths) and corresponding DARTS cell (right). Compared to a standard UNet, the network displays a dense connectivity at higher spatial resolution.}
\label{fig:UNet++_decoded}
\end{figure}

\paragraph{}As presented in Figure~\ref{fig:UNet++_decoded}, the decoded network exhibits very dense connectivity at maximum spatial resolution and retains the usual U-Net high semantic path. This intuitively makes sense: to perform proper spatial segmentation, processing full resolution patches is critical while the classification task can benefit from higher semantic content. The decoded cell displays a variety of operations with both convolution layers and pooling/skip layers. For this experiment, only classical convolutions were ultimately chosen however.

\begin{figure}[!ht]
    \centering
    \includegraphics[scale=0.3]{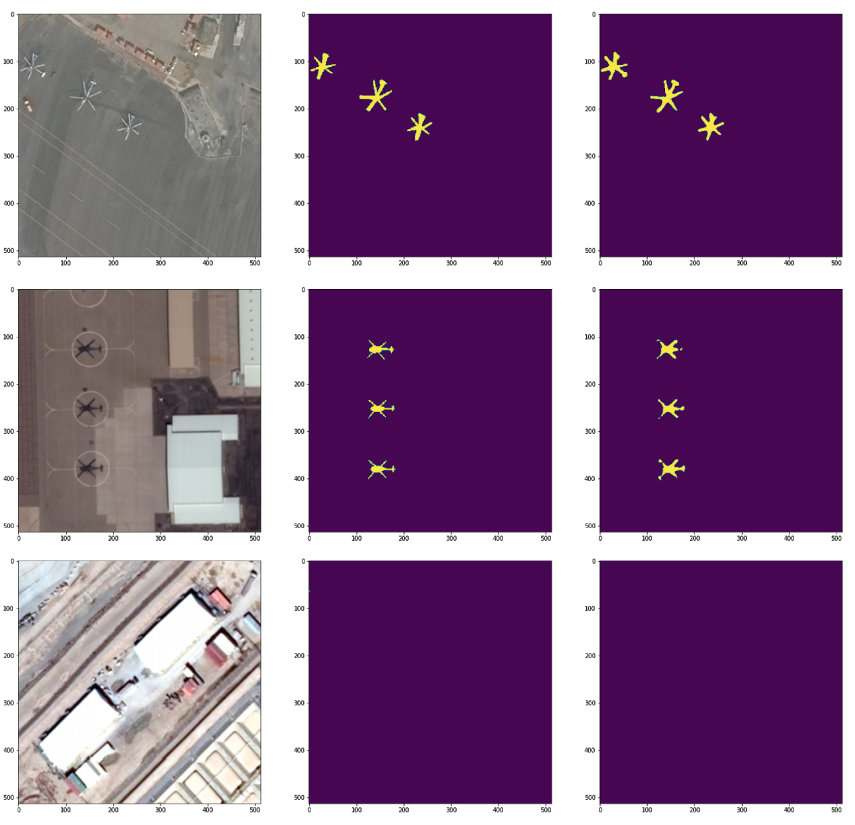}
    \caption{From left to right: image, ground truth and DARTS-UNet++ predictions.}
    \label{fig:darts_unet++_preds}
\end{figure}

\paragraph{}On a qualitative basis -- see Figure~\ref{fig:darts_unet++_preds} -- the performed segmentations delineate quite well the shapes of the detected helicopters. However, from a quantitative standpoint, this architecture didn't manage to outperform either our baseline or the structure found using the ResNeXt-UNet topology. 

\subsection{Random search baselines}

\paragraph{}Evaluating the potential of a new method always requires to compare against a competitive alternative. Following recommendations made on NAS experiment reproducibility \cite{nasrandom}, we decided to compare our best searched DARTS cells against random search. To this end, based on the DARTS-UNet and ResNeXt-UNet cell topologies, we apply a random selection of each node's operation by uniformly sampling with replacement. We call these random cells as baselines in our experimental protocol.

\begin{figure}[!ht]
\centering
\begin{subfigure}{.45\textwidth}
  \centering
  \includegraphics[width=1\linewidth]{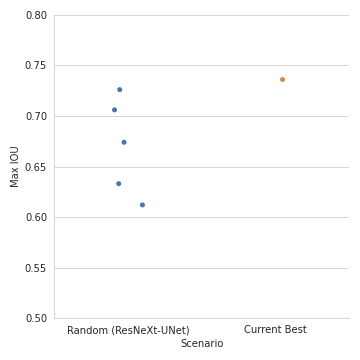}
  \caption{ResNeXt-UNet cell}
  \label{fig:random_resnext}
\end{subfigure}%
\begin{subfigure}{.45\textwidth}
  \centering
  \includegraphics[width=1\linewidth]{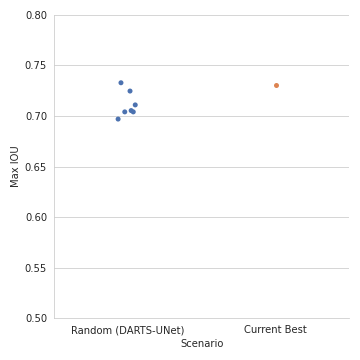}
  \caption{DARTS-UNet cell}
  \label{fig:random_darts_unet}
\end{subfigure}
\caption{Performance comparison of random search and DARTS for two types of cell topology.}
\label{fig:random_baseline}
\end{figure}

\paragraph{}Random search is a simple alternative to DARTS, though it is still necessary to train the sampled cell in the same way as before to obtain a fair comparison. In figure \ref{fig:random_baseline} we illustrate the maximum IoU obtained after the training of the selected cells (through DARTS or random search). We experimentally observed that random search can reach performances similar to DARTS, thus using DARTS-based NAS on this specific use case amounts to finding a best performing cell with random search.

\section{Conclusions and feedback}\label{sec:conclusions}

\paragraph{}As mentioned in this paper opening, the aim of this first study was threefold. NAS being presented as an interesting network design tool in the literature, our first intent was to find the most relevant methods to build such a tool and to implement a framework based on it. Upon building this tool, our second objective was to put it to use on one of our company use-case and benchmark whether or not more suitable architectures could be found using it. Finally, and following these experiments, we wanted to present some first conclusions and use our feedback to propose new ideas to build on this study.

\paragraph{}In terms of theory and method, NAS is still a very young subject, thus lacking stability and in depth insights. When compared to classical deep learning, the number of hyper-parameters drastically increases, and when considering the continuous relaxation framework, the joint optimization to perform is vastly more complex. A striking example of how this complexity translates to experiment instability: the high variability in terms of network structure and the corresponding performance between two similar searches. Indeed the probed architectures are initialized at random -- as we obviously can't rely on usual ImageNet weights -- and therefore suffers from their higher number of trained parameters. This lack of maturity is also exhibited by the numerous studies arguing that random search may have similar performances -- or better ones -- given similar search time. This is one of the main conclusions of this study, which wasn't especially expected when considering performances displayed by the technology in some of the NAS key papers.

\paragraph{}As noted in the literature, the computing power required to use such techniques is still a very limiting factor -- even when considering recent techniques to reduce it. Not every company can afford ownership or access to an HPC infrastructure. Moreover, implementing a framework running properly in such multi-nodes/multi-GPUs context is not trivial. Details are presented in Appendix \ref{sec:implementation}, but in a nutshell, one ends up using very experimental parts of relevant backend and libraries and therefore encounters rare cases of failures and unexpected behaviours -- for which only a very limited community may be able to help.

\paragraph{}Applying NAS to one of our use-cases, we obtained mixed results. On the one hand, when challenging an heavily tailored architecture, we did not manage to find an outperforming structure. Multiple reasons exist for this outcome -- e.g. we searched for a single cell structure when the reference architecture possess various tailored ones. On the other hand, when considering the sheer time involved in designing our production architecture, NAS managed to find a competitive structure much faster and at a far lower human-involvement cost -- upon completion of the framework implementation. Taking into account the possible biases in our helicopter experiment, we benchmarked NAS against a less use-case tailored architecture. Preliminary results are more encouraging: the NAS-found structures manage to reach the same level of performance -- and even to outperform it. However, we would not use NAS to design a better NN for this use-case.

\paragraph{}By conducting this first study, we managed to get a better grasp on the core theory and technology involved in NAS and implement a first framework enabling us to conduct our experiments. Following these results, further studies will focus on potentially missing tools to our framework or better conditioning of our search strategy. As already noted, a large randomness is introduced in our probed super-network structures due to the increased number of parameters and inability to rely on pre-trained weights. As such, a tool to assess structural soundness, previous to any search / training phase, would be a great asset in NAS. Taking inspiration on recent propositions \cite{mellor2021neural, chen2021neural, lin2021zennas}, one could investigate criteria such as \textit{expressivity} and \textit{trainability} to build such a metric. Regarding the high complexity involved when searching for an entire network / cell structure, one could try to reduce it and embed it in an already well designed NN. As an example, in an handcrafted network, one could try to improve the global performance by NAS-searching for a very limited subsection of the network -- e.g. at custom layer level. A question then arises: how to select the blocks that need improvement ? During search steps, we noticed that some large variation can exist between $\alpha$ entropy levels of the different cells composing the network. As such, the corresponding gradient applied during back-propagation could be more or less important, and our intuition is that such a criteria could be used to estimate which part of the network will benefit more from a change in its design.

\paragraph{}Following this first study, it is indeed our intuition that one needs to better condition the propositions made in the literature to enable a really suitable use of NAS in industry applications.

\renewcommand{\abstractname}{Acknowledgements}
\begin{abstract}
 This work was granted access to the HPC resources of IDRIS under the allocation 2020-GC-101472 made by GENCI.
\end{abstract}

\printbibliography
\clearpage
\appendix

\section{Implementation for HPC use}\label{sec:implementation}

\paragraph{}Thanks to the \textit{Grand Challenge Jean-Zay} program, initiated by the GENCI, we were allocated computing hours on their HPC infrastructure. This enabled us to conduct NAS experiments that would otherwise not have been possible on our proprietary hardware. In terms of computing power, the partition of the cluster opened to us owned 351 nodes each equipped with 4 GPUs NVIDIA V100, 2 Intel cascade processors and 192 Gb of RAM. 
\paragraph{}To fully take advantage of these hardware capabilities, our framework needed to properly handle both model parallelism (discrete approach) and data parallelism (continuous approach). In the following, we will develop on the tools we used and the issues we encountered during our implementation.

\subsection{Model parallelism: Mpi4Py}
\paragraph{}Although not mentioned in this paper, we conducted experiments using evolutionary NAS (E-NAS). To do so, we chose to implement the search using a multiprocessing dispatch approach:
\begin{itemize}
    \item a parent process acting as an orchestrator
    \item children processes performing training / evaluation
\end{itemize}

\noindent By \textit{orchestrator}, we here mean that the process handles the architectures population evolution -- of course, to prevent idling, it also performs training tasks. The Message Passing Interface (MPI)\cite{MPI} is a library widely used on HPC infrastructures. It was the perfect dispatching tool for our orchestrator and, as our framework was coded in Python, we relied on the dedicated package MPI4Py \cite{MPI4py}.

\paragraph{} For clarity purpose, the MPI dispatching pseudo-code is presented in Algorithm \ref{alg:discrete}. In this implementation, upon architectures dispatch to all GPUs by the orchestrator, we ensure that each device is indeed used and training a specific NN. By nature, this code section is asynchronous: although computations are performed on an homogeneous infrastructure, each NN architecture will need a different time to be trained on a specific number of epochs -- relying heavily on how complex the probed NN architecture is. Synchronisation therefore is ensured by the orchestrator: computations are dispatched by it and, upon completing its own training, it awaits a signal from each GPU providing the results of its computation. This is easily implemented using MPI4Py through devices peer-to-peer communications. Relying on Tensorflow to perform said trainings, we encountered some unexpected issues. 

\paragraph{} The major one lies in the high variability in terms of VRAM consumption for each architecture probed. As direct consequence, upon not being able to estimate the consumption of the biggest architecture possible, one will be subject to Out Of Memory (OOM) errors that will perturb the search. However, it is not trivial to evaluate the precise maximum VRAM consumption for the search, even when you can roughly estimate the structure of your most complex network -- this is mainly linked to the graph abstraction used by Tensorflow to encapsulate all part of the training process. Big providers of cloud DL platform have identified this issue and are currently studying it \cite{gao2020estimating}.

\paragraph{} The second one is of a more nitpicking nature, but if not anticipated, one may incur a major code refactoring. In any Tensorflow backed training framework, the most commonly found import is the generic \texttt{import tensorflow as tf}. However, following this import, libraries opening are performed -- e.g. CUDNN. In the context of a multiprocessing dispatching of training tasks on a multi-node multi-GPUs infrastructure, this may cause conflicts that terminate code execution. In a nutshell, for the specific case of dispatching previously explained, this could very simply be avoided: the codebase should be structured in such a way that any \texttt{tensorflow import} is strictly performed at child process level. In this way, you can ensure that the task has been submitted properly -- through MPI -- to a dedicated GPU and can therefore proceed as any vanilla training perform on a single GPU computer.\\

\begin{algorithm}[H]
\label{alg:discrete}
\SetAlgoLined
\KwResult{Hist. storing arch. and corresponding performances}
 ${Orch. \leftarrow GPU intel}$\;
 ${Orch. \underset{init.}{\rightarrow} (Pop., Hist.)}$\;
 \While{Epoch}{
     \For{(i, arch.) $\in$ enumerate(population -- 1)}{
        ${Orch. \underset{arch}{\rightarrow} GPU_{i}}$\;
        }
    ${Train(arch)\vert_{Orch. \cup GPUs}}$\;
     \For{(i, arch.) $\in$ enumerate(population -- 1)}{
     ${Orch. \underset{arch, perf}{\leftarrow} GPU_{i}}$\;
     ${Orch. \underset{arch, perf}{\rightarrow} Hist.}$\;
     }
    ${Evolve(Pop.)\vert_{Orch.}}$\;
    }
\caption{MPI communication (discrete framework)}
\end{algorithm}

\subsection{Data parallelism: TensorFlow and Horovod}

\paragraph{}As training heavy search networks requires a significant amount of computing resources, we rely on data parallelism to keep the training time of such networks reasonable. In such a setting, two cases are to be considered:

\begin{itemize}
    \item the usage of multiple GPUs on a single node, that we denote \textit{mono-node/multi-gpu}
    \item the usage of multiple GPUs on multiple nodes, that we denote \textit{multi-node/multi-gpu}
\end{itemize}

\noindent
We use TensorFlow \cite{tensorflow} as our reference deep learning framework, and so consider the native \textit{mirrored strategy} available in the package. For our two use-cases, two distinct classes exist, respectively \textit{MirroredStrategy} and \textit{MultiWorkerMirroredStrategy}. Although we encountered no issue with the former, such a thing cannot be said for the latter.

\paragraph{}We follow user guidelines and establish a generic codebase that handles both cases, based on \textit{MultiWorkerMirroredStrategy}. As the majority of the methods we use are contained in an \textit{experimental} package, and we are building a custom training loop, it seems reasonable to expect potential issues. As discussed before, while we encountered no difficulty working with a single node, it was not the case for multiple nodes. Most of the time, multi-node/multi-gpu trainings were stuck on a training or an evaluation step before reaching the end, most likely while applying reduction on all partial results from each worker, which implies a communication error between nodes. Such issues are also hard to solve as we do not get any feedback that can help identify the point of failure.

\paragraph{}In a second step, we considered Horovod \cite{horovod}. As we did not manage to solve the encountered TensorFlow issues, we relied on a dedicated and proven framework, widely used for distributed computing on HPC. Albeit this framework is harder to use initially -- as any user needs to understand basic MPI concepts -- it remains easy to integrate in an existing codebase. We updated our codebase to replace TensorFlow's mirrored strategy by Horovod. As the latter explicitly uses MPI concepts that are also used implicitly in the former, it is easier to understand and follow the training process. We encountered no blocking issue with Horovod for both use-cases, except some unusual behaviours (e.g deadlocks or crashes) that occurred only once. Additionally, switching to Horovod provided a small performance gain in terms of training time. Building upon these observations, we chose Horovod as our reference framework for data parallelism.

\end{document}